\documentclass[10pt,twocolumn,letterpaper]{article}

\usepackage[pagenumbers]{cvpr} 

\newcommand{\projectname}{\textsc{RoboBuilder}\xspace}

\usepackage{microtype}

\usepackage{booktabs,multirow,xcolor,graphicx}
\usepackage{graphicx,tcolorbox,enumitem}
\usepackage{booktabs,makecell,tabularx,array}
\usepackage{colortbl}
\usepackage{adjustbox}
\usepackage{placeins}

\definecolor{bestscoregreen}{RGB}{185,222,180}
\definecolor{secondscoregreen}{RGB}{228,242,223}
\newcommand{\bestscore}[1]{\cellcolor{bestscoregreen}\textbf{#1}}
\newcommand{\secondscore}[1]{\cellcolor{secondscoregreen}#1}

\definecolor{cvprblue}{rgb}{0.21,0.49,0.74}
\usepackage[pagebackref,breaklinks,colorlinks,allcolors=cvprblue]{hyperref}

\def\confName{CVPR}
\def\confYear{2027}

\title{RoboBuilder: A Precision Assembly Benchmark for Robot Foundation Models}

\author{%
\parbox{0.95\textwidth}{\centering
Vineet~Bhat$^{1}$ \quad
Mikaela~Angelina~Uy$^{2}$ \quad
Alex~Zook$^{2}$ \quad
Chan~Hee~Song$^{2}$ \quad
Hugo~Hadfield$^{2}$ \quad
Prashanth~Krishnamurthy$^{1}$ \quad
Ramesh~Karri$^{1}$ \quad
Jenai~Xuning~Yang$^{2}$ \quad
Stan~Birchfield$^{2}$ \quad
Valts~Blukis$^{2}$ \quad
Farshad~Khorrami$^{1}$ \quad
Jonathan~Tremblay$^{2}$
\\[6pt]
$^{1}$New York University \qquad $^{2}$NVIDIA
}}

\begin{document}
\maketitle

\begin{abstract}

Robotics is currently undergoing a paradigm shift as recent foundation models enable zero-shot control of robots for manipulation tasks. 
Current benchmarks for evaluating robot policies ignore high precision and long horizon assembly tasks and 
focus on computing a binary success rate based on loosely defined goals, such as, placing an object on a centimeter scale goal receptacle region. 
We introduce RoboBuilder, a suite of short- and long-horizon tasks featuring a diverse collection of assets drawn from the mechanical assembly literature. 
These tasks span varying levels of manipulation difficulty, requiring robot policies to achieve relative, metric, and constrained placements with millimeter precision. 
Tasks in RoboBuilder reflect real-world manufacturing challenges and are designed to represent real world applications, where goal specifications are both described by text and/or CAD diagrams. 
We show that powerful models like GPT~6~Astra do not achieve a high level of accuracy, reporting 1.8\% success at 1 mm thresholds on \projectname-Short demonstrating the need for further research on precision manipulation. 
On \projectname-Long, which tests model capacity to assemble simple objects, GPT~6~Astra with CAD inputs cannot build any of the assemblies within 2\,mm and only one assembly at 10\,mm. 
A qualitative real-world study on two 3D-printed assemblies shows that these precision failures carry over to physical robots.
We will release our benchmark tasks and metrics to boost community-driven research in improving the precision of future robot policies: \url{https://vineet2104.github.io/RoboBuilder/}
\end{abstract}

\section{Introduction}
\label{sec:intro}

\begin{figure*}[!t]
    \centering
    \includegraphics[width=\textwidth]{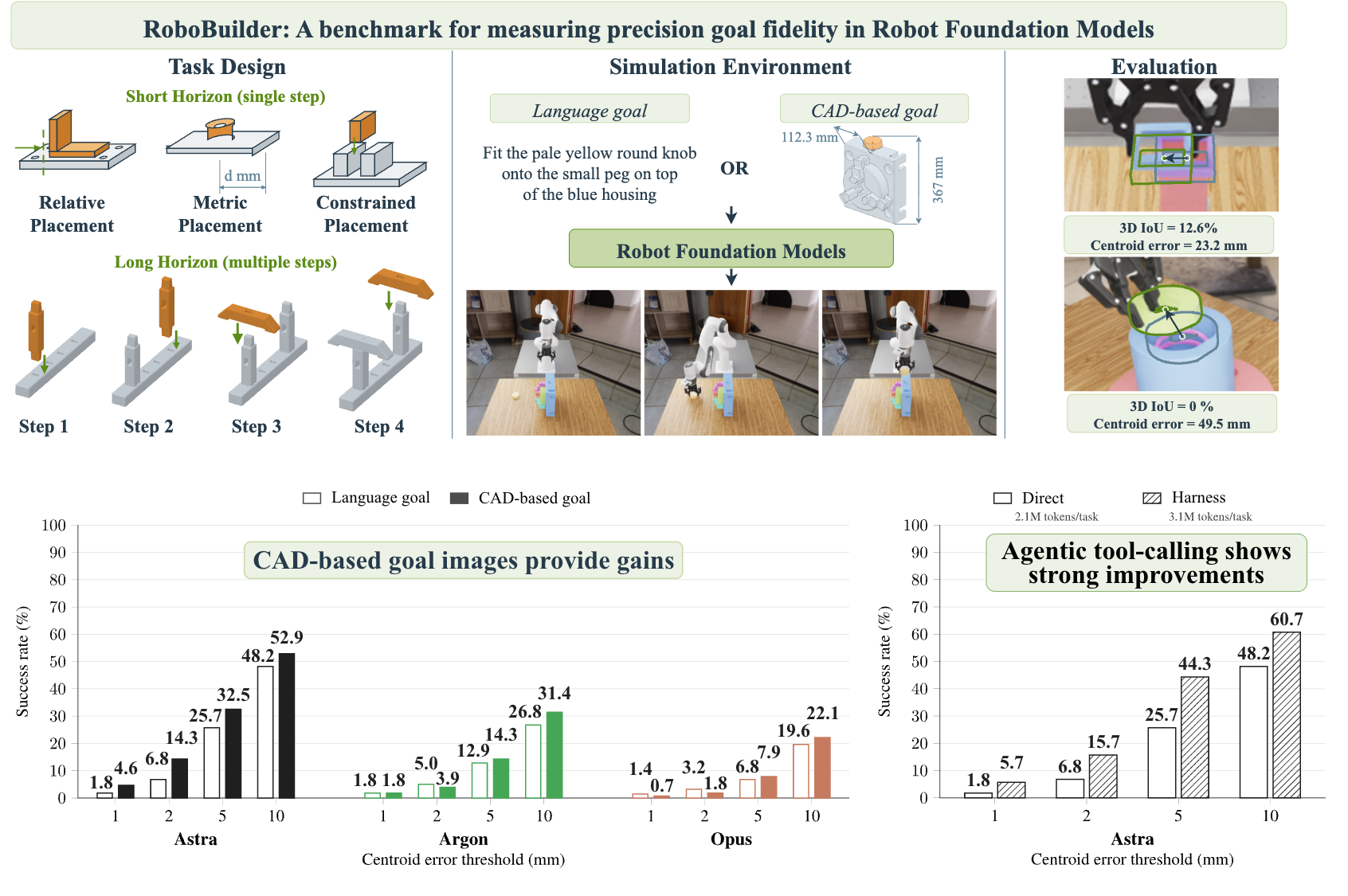}
    \caption{
    RoboBuilder measures placement precision overlooked by existing benchmarks. 
    Across short- and long-horizon assembly tasks, current robot foundation models struggle to achieve millimeter-level accuracy. 
    CAD-based goals improve precision, but substantial errors remain.
    }
    \label{fig:teaser}
\end{figure*}

\vspace{-0.2cm}
Mechanical assembly is an important application of robotics in manufacturing, requiring accurate placement and contact-rich manipulation~\cite{Narang-RSS-22,industreal,Tang-RSS-24}.
Parts are typically arranged on a workspace, with a task description, visual instructions, or both specifying the target assembly~\cite{wangikea,Zhang_2025_ICCV}. 
Completing the task requires identifying each part, determining how the parts fit together, and placing them precisely~\cite{Heo-RSS-23}.


Robot foundation models (RFMs), including vision-language-action models (VLAs)~\cite{bjorck2025gr00t,pmlr-v305-black25a,rt2,openvla}, world action models (WAMs)~\cite{nvidia2026cosmos3}, and large language models (LLMs) applied to robot control~\cite{chen2026volo,codeaspolicies,voxposer}, have made progress in object manipulation. 
For example, Astranomy~\cite{yang2026astranomy} reports approximately 75\% success on the RoboLab-120 benchmark~\cite{yang2026robolab} with GPT 6 Astra.
Yet task success alone does not establish whether these models can perform precise assembly. 
Benchmarks including LIBERO~\cite{libero}, RoboTwin~\cite{chen2025robotwin}, RoboDojo~\cite{chen2026robodojo} , and CALVIN~\cite{calvin} 
use task-specific success criteria, including geometric checks
such as reaching within a distance threshold, placing an object
inside a container, or establishing contact with another object.
Measuring progress toward general assembly requires evaluating both task completion and the accuracy of the resulting assembly.


To address this gap we introduce \projectname, a robotics benchmark and evaluation protocol that measures millimeter-scale goal following capabilities in RFM policies.
\projectname has 295 tasks: 280 short-horizon single-step tasks (\projectname-Short) and 15 long-horizon multi-step tasks (\projectname-Long) all using up to four cameras and a Franka robot.
Each task in \projectname-Short requires the robot to grasp a target part and place it at a specified pose relative to a fixed receiving assembly.
We evaluate how closely each placed part matches its target configuration using two complementary measures. 
The centroid error measures the distance between the part's achieved center and its target center.
We report the percentage of tasks within 1, 2, 5, and 10\,mm as positional success rate.
3D intersection over union (3D IoU) measures the overlap between the part's goal and achieved poses using their shared volume divided by the total volume occupied by either configuration.

Our results show that evaluated RFMs struggle with precise object placement. 
When prompted directly with language goals, GPT~6~Astra, the strongest evaluated model in this setting, achieves a positional success rate of \textbf{$\textbf{48.2\%}$ at 10\,mm}, decreasing to \textbf{$\textbf{1.8\%}$ at 1\,mm}.
We compare two interventions: replacing language goals with CAD-based goals of the target configuration, and providing GPT~6~Astra with a Codex harness and access to writing perception and computational tools, together with a larger token allowance.
For GPT~6~Astra, CAD-based goals increase positional success rate at 1/2\,mm from $1.8 / 6.8\%$ to $4.6 / 14.3\%$.
GPT~6~Astra with language goals and Codex harness reaches $5.7/15.7\%$.
3D IoU increases from $29.6\%$ with language goals and direct inference to $35.1\%$ with CAD-based goals and $40.4\%$ with language goals and the Codex harness.
Our more challenging benchmark, \projectname-Long, contains 15 assemblies, including chairs, stools, tables, a shelf, and a beam, each requiring 2--9 steps.
With CAD-based goals and the Codex harness, GPT~6~Astra places every required part within 10\,mm of its target center in just one of the 15 assemblies.

Our contributions are:

\begin{enumerate}
    \item \projectname-Short, a collection of 280 single-step assembly tasks. We evaluate positional success rate at centroid-error thresholds of 1, 2, 5, and 10\,mm and mean target-part 3D IoU.
    \item Evaluations of current RFMs with language goals, CAD-based image goals, and an agentic harness through Codex. The evaluated models achieve low positional success rates at 1\,mm and 2\,mm, with substantial room for improvement even at 10\,mm (Figure~\ref{fig:teaser}).
    \item \projectname-Long, a collection of 15 multi-step assembly tasks. With CAD-based goals and the Codex harness GPT~6~Astra places every required part within 10\,mm of its target center in only one assembly, Mammut-table, with a mean per-part 3D IoU of $72.9\%$.
\end{enumerate}

\section{Related Work}
\label{sec:related-works}
\vspace{-0.2cm}

\noindent\textbf{Robotic assembly.}
Mechanical assembly has long motivated research on precise, contact-rich robotic manipulation~\cite{kimble2020nist}. 
Factory~\cite{Narang-RSS-22} provides efficient simulation and reinforcement learning environments for assembly, and IndustReal~\cite{industreal} transfers simulation-trained assembly skills to physical robots.
AutoMate~\cite{Tang-RSS-24} learns specialist and generalist policies across diverse part geometries, while FORGE~\cite{forge} addresses force-aware execution under pose uncertainty. Assemble Them All~\cite{assemblethemall} and ASAP~\cite{asap} plan assembly sequences and part motions.
Previous works have also showed that CAD geometry supports planning and control, such as using geometric motion plans to guide learned controllers~\cite{assemblyfromcad}. 
Works such as Fabrica~\cite{tian2025fabrica} integrate assembly planning, grasp
and fixture generation, and reinforcement learning for multi-part
robotic assembly, while FurnitureBench~\cite{Heo-RSS-23} provides real-world and simulated environments for long-horizon assembly, and LABIT~\cite{labit} focuses on industrial assembly tasks.
These works provide assets, planning methods, assembly skills,
and evaluation environments. RoboBuilder complements these
works with an evaluation protocol for generalist robot
policies separating procedural assembly progress from
geometric fidelity of final placements under language and CAD-based goals.\\
\noindent\textbf{CAD and geometry instructions for assembly.}
Assembly goals must specify the intended geometry and
relationships of the parts. 
AutoMate~\cite{cad_automate} and JoinABLe~\cite{joinable} study these relationships by learning to predict mates between CAD parts. 
Visual assembly instructions offer a way to communicate assembly procedures~\cite{assemblyinstructions}.
Instruction-guided approaches build on resources such as IKEA-Manual~\cite{wangikea}: Manual-PA~\cite{Zhang_2025_ICCV} predicts part configurations from diagrams, while Manual2Skill~\cite{tie2025manual} connects manual interpretation with pose estimation and robotic execution.
AssemblyBench~\cite{Li_2026_CVPR} extends this direction to industrial objects and physically plausible part trajectories.
Other existing works examine the interpretation of projected
geometry: SPARE3D~\cite{spare3d} studies spatial reasoning over three-view line drawings, PlankAssembly~\cite{plankassembly} and CAD2Program~\cite{cad2program} reconstruct 3D models from vector and raster drawings, respectively, and MechVQA~\cite{mechvqa} evaluates mechanical drawing understanding by multimodal models. 
RoboBuilder builds on these lines of works by using dimensioned projected CAD diagrams as explicit goals for robot execution. 
We compare language and CAD goal specifications using the same geometric metrics to evaluate the fidelity of the resulting part placements.\\
\noindent\textbf{Simulation benchmarks for robot learning.}
Benchmarks such as LIBERO~\cite{libero}, RoboTwin 2.0~\cite{chen2025robotwin}, RoboCasa365~\cite{robocasa365}, CALVIN~\cite{calvin}, RLBench~\cite{rlbench}, and ManiSkill2~\cite{maniskill2} support systematic evaluation of instruction following, manipulation, and generalization across objects and environments. 
Their tasks include pick-and-place, articulated-object interaction, and multi-step manipulation. 
RoboDojo~\cite{chen2026robodojo} evaluates precision through tasks with strict spatial requirements and reports task success and partial progress without quantification.
These evaluations motivate complementary measurements of positional accuracy and volumetric overlap.
Table~1 compares task coverage and reported measures across selected benchmarks, including RoboLab~\cite{yang2026robolab} and BEHAVIOR-1K~\cite{li2024behavior1k}. 
It uses a placement example to show how selected task checks differ from measurements of position accuracy and target overlap. 
Each task can require more conditions. 
A passed check does not report the remaining placement error. 
Beyond task-level success criteria, SIMPLER~\cite{simpler}
studies the fidelity of simulation-based policy evaluation
through paired simulation--real comparisons.
RoboBuilder complements these benchmarks with precise placement tasks specified through language goals and CAD-based goals. 
We evaluate placement fidelity using positional success rate at centroid-error thresholds of 1, 2, 5, and 10\,mm and mean 3D IoU. 
For multi-step assemblies, we separately report procedural assembly progress and the number of completed steps that meet each positional tolerance.

\begin{table}[!t]
\centering
\newcommand{\rbtablepass}{\textcolor[rgb]{0,0.45,0.15}{\ensuremath{\checkmark}}}
\newcommand{\rbtablefail}{\textcolor[rgb]{0.75,0.08,0.08}{\ensuremath{\times}}}
\caption{
Illustration of success checking in various benchmarks.
A 20\,mm cube rests on a surface, 8\,mm from a marked target.
The $y$, $z$, and rotation errors are zero.
The results illustrate selected success condition checking. 
SR = success rate. IoU = intersection over union.
The symbols \rbtablepass{} and \rbtablefail{} show whether the example passes or fails the selected benchmark check, respectively. \vspace{-0.2cm}
}
\label{tab:benchmark_comparison}
\footnotesize
\setlength{\tabcolsep}{2pt}
\renewcommand{\arraystretch}{1.15}
\begin{minipage}[c]{0.33\columnwidth}
\centering
\setlength{\unitlength}{0.26mm}
\begin{picture}(100,65)
\thinlines
\put(1,8){\framebox(98,55){}}
\put(7,53){\makebox(0,0)[l]{\footnotesize Top view}}
\put(32,24){\dashbox{1}(20,20){}}
\linethickness{0.8pt}
\put(40,24){\framebox(20,20){}}
\thinlines
\put(32,0){\makebox(0,0)[r]{\footnotesize Target}}
\put(61,0){\makebox(0,0)[l]{\footnotesize Final}}
\end{picture}
\end{minipage}\hfill
\begin{minipage}[c]{0.65\columnwidth}
\textbf{Running example}\par
The object touches the surface.\par
Centroid error: 8\,mm\par
3D IoU: 42.9\%\par
Dashed: target. Solid: final.
\end{minipage}
\par\vspace{6pt}
\begin{tabularx}{\columnwidth}{@{}
>{\raggedright\arraybackslash}p{0.31\columnwidth}
>{\raggedright\arraybackslash}p{0.21\columnwidth}
>{\raggedright\arraybackslash}X@{}}
\toprule
\textbf{Benchmark} & \textbf{Results} & \textbf{Example check and result} \\
\midrule
\mbox{\href{https://arxiv.org/pdf/2112.03227}{CALVIN}~\cite{calvin}}\newline (34 tasks)
& Task SR, chained SR
& \href{https://github.com/mees/calvin_env/blob/1431a46bd36bde5903fb6345e68b5ccc30def666/calvin_env/envs/tasks.py\#L158-L198}{Object in slider/drawer}\newline \rbtablepass\ Contact check \\
\addlinespace[2pt]
\mbox{\href{https://arxiv.org/pdf/2306.03310}{LIBERO}~\cite{libero}}\newline (130 tasks)
& Task SR, transfer scores
& \href{https://github.com/Lifelong-Robot-Learning/LIBERO/blob/8f1084e3132a39270c3a13ebe37270a43ece2a01/libero/libero/envs/predicates/base_predicates.py\#L65}{Object on the surface}\newline \rbtablepass\ Relation holds \\
\addlinespace[2pt]
\mbox{\href{https://arxiv.org/pdf/2506.18088}{RoboTwin 2.0}~\cite{chen2025robotwin}}\newline (50 tasks)
& Task SR
& \href{https://github.com/RoboTwin-Platform/RoboTwin/blob/ea8b21121ebb3cd201ff5b3fe361944ac94eda3f/envs/place_mouse_pad.py\#L109}{$|e_x|<15$, $|e_y|<12$\,mm}\newline \rbtablepass\ Position check \\
\addlinespace[2pt]
\mbox{\href{https://arxiv.org/pdf/2603.04356}{RoboCasa365}~\cite{robocasa365}}\newline (50 target tasks)
& Task SR
& \href{https://github.com/robocasa/robocasa/blob/425684d332f135f5b30a40b321f52d15a0fe9250/robocasa/utils/object_utils.py\#L621}{Surface contact helper}\newline \rbtablepass\ Contact check \\
\addlinespace[2pt]
\mbox{\href{https://arxiv.org/pdf/2607.04434}{RoboDojo}~\cite{chen2026robodojo}}\newline (60 tasks)
& Task SR, partial score
& \href{https://github.com/RoboDojo-Benchmark/RoboDojo/blob/f57e1633ffed59761bc8e568cfe7530c4b33381c/task/RoboDojo/tasks/play_stacking_toy.py\#L21}{Peg alignment $<1$\,mm}\newline \rbtablefail\ Alignment check \\
\addlinespace[2pt]
\mbox{\href{https://arxiv.org/pdf/2604.09860}{RoboLab-120}~\cite{yang2026robolab}}\newline (120 tasks)
& Task SR, subtask score, motion quality
& \href{https://github.com/NVlabs/RoboLab/blob/8a4f58102a2a4468f108ae61369e9248b79ed426/robolab/core/task/conditionals.py\#L193}{Object relation predicates}\newline \rbtablepass\ Relation holds \\
\addlinespace[2pt]
\mbox{\href{https://behavior.stanford.edu/challenge/}{BEHAVIOR-1K}~\cite{li2024behavior1k}}\newline (100 tasks)
& Goal score, efficiency
& \href{https://arxiv.org/html/2403.09227\#A5.SS1.SSS4}{Above the surface with contact}\newline \rbtablepass\ One goal condition \\
\midrule
\textbf{\projectname}\newline (295 tasks)
& \textbf{Positional SR}\newline \textbf{Mean 3D IoU}\newline Assembly progress
& \rbtablefail\ \textbf{1, 2, 5\,mm}\newline \rbtablepass\ \textbf{10\,mm}\newline \textbf{3D IoU: 42.9\%} \\
\bottomrule
\end{tabularx}
\vspace{-0.5cm}
\end{table}

\section{Methodology}
\label{sec:method}

We describe the robot, simulation environment, task construction and verification, language and CAD-based goals, and metrics for placement fidelity and assembly progress.

\subsection{Environment Setup}
\label{subsec:environment}

We use a Franka Panda arm with a Robotiq-2f-85 gripper in the RoboLab tabletop simulation environment~\cite{yang2026robolab}. 
We use a DROID-inspired observation setup~\cite{droid} with
two shoulder cameras and one wrist camera.
We provide an additional front-facing camera view to the LLM baselines.
Assets for each task are initialized on the tabletop within the robot's workspace.
The robot controller supports both absolute joint-position control and end-effector control.

\subsection{Task Creation}
\label{subsec:task_creation}

We construct benchmark tasks using 3D models from Fusion360~\cite{joinable} and IKEA-Manual~\cite{wangikea}, which provide manufacturing-relevant assemblies with individual parts and final assembled configurations. 
We instantiate assemblies with separate part geometries and relative poses, enabling independent manipulation and evaluation against each target pose.
For each imported assembly, we run the ASAP sequence planner~\cite{asap} to obtain 3D trajectories for part placement or insertion. 
We discard assemblies for which ASAP does not return a valid plan within 600\,s.

\begin{figure*}[!ht]
    \centering
    \includegraphics[width=\textwidth]{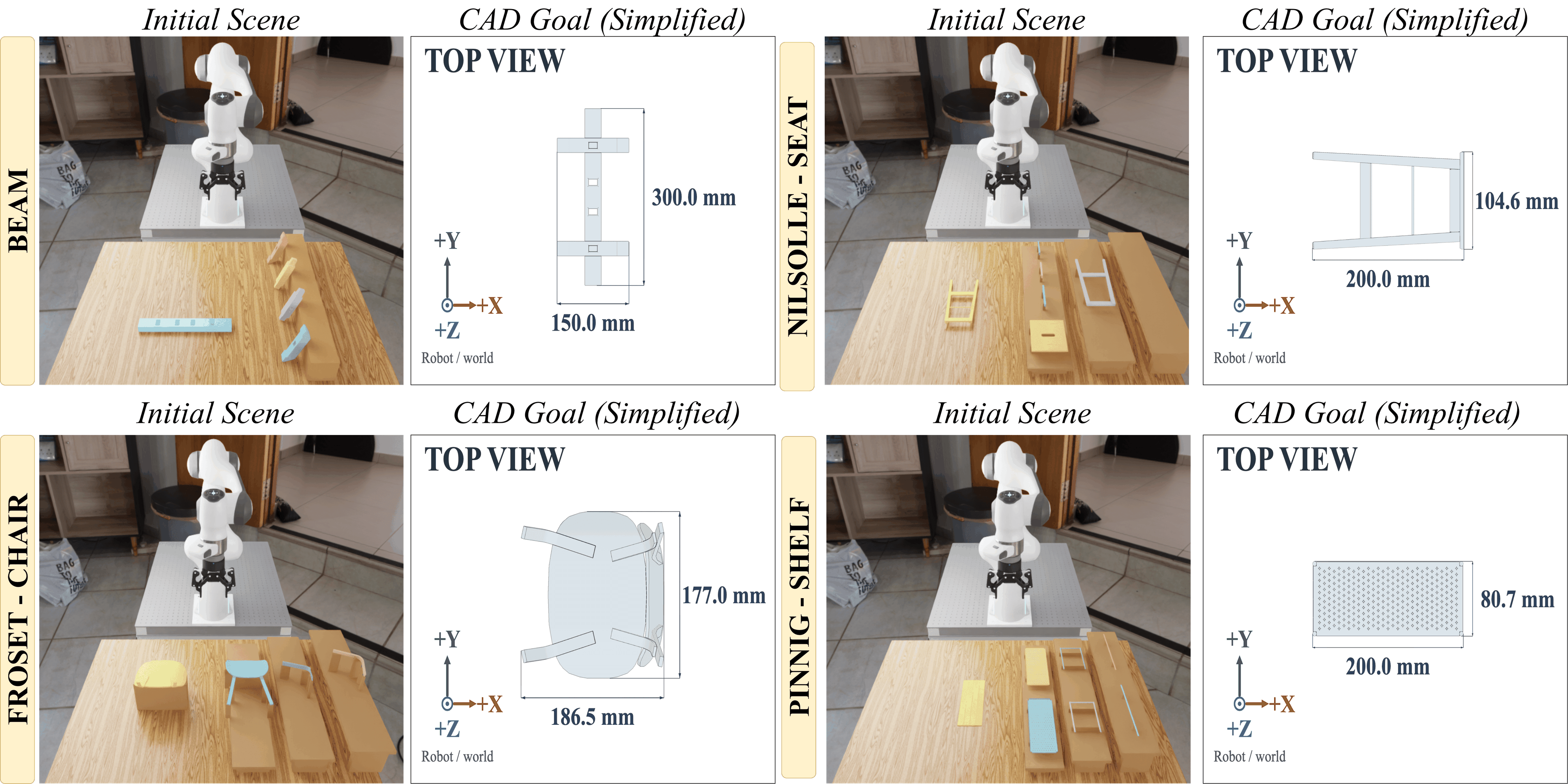}
    \caption{
     Examples of initial scenes and simplified CAD-based goal diagrams for tasks in \projectname-Long. The initial scene shows the qualified scene configuration after verification protocol (Sec~\ref{subsec:verification}). The CAD-based goal on the right of each scene describes the intended assembly with dimensions and robot reference orientation. Expanded views of CAD-based goal diagrams are in the supplementary material.
    }
    \vspace{-10pt}
    \label{fig:long_horizon_examples}
\end{figure*}

\subsection{Task Verification}
\label{subsec:verification}

A geometrically feasible assembly plan requires grasps and motions that the robot can execute.
We verify each benchmark task by constructing an oracle part trajectory, converting it into an end-effector trajectory under a stable grasp, and checking whether the robot can follow that trajectory.

This process begins with randomly initializing the assembly parts on the tabletop.
For each assembly step, a fixed waypoint controller constructs a trajectory that lifts the target part from its initial pose, transfers it to the start of the corresponding ASAP~\cite{asap} trajectory, and follows that trajectory for the final placement or insertion. 
We sample 6 DoF grasps and keep the part-to-gripper transform fixed for each candidate grasp.
We use this transform to convert the oracle part trajectory into end-effector poses. 
Inverse kinematics then produces joint configurations along the trajectory, subject to joint limits and the solver's pose tolerances.
We retain a candidate only when the grasp and resulting robot motion plan pass our verification procedure.

When an initial scene fails verification, we resample the initial part poses and repeat the procedure, allowing up to five initialization attempts. 
We exclude steps for which we find no valid motion plan. 
For \projectname-Long, we retain assemblies only when every step passes verification sequentially from a common initial scene (see supplementary material).
We retain individually verified steps from otherwise unsuccessful assembly sequences as tasks in \projectname-Short, together with their verified initial scenes.
The resulting benchmark contains 280 single-step tasks in \projectname-Short and 15 multi-step assemblies in \projectname-Long.

\subsection{Goal Annotation}
\label{subsec:goal_annotation}


\paragraph{CAD-based goals.}
For each task, we construct diagrams of the target configuration 3D model from four views: front, right, top, and isometric. 
We generate the images using a custom Python renderer built by computing part and relative placement dimensions directly from the verified assembly geometry and include them in the diagrams. 
All four views use orthographic projection, with a fixed oblique viewing direction for the isometric view; a depth buffer determines which mesh surfaces are visible. Tasks in \projectname-Long are only evaluated through CAD-based goals to provide the robot policy with a concise yet rich context surround part dimensions and mating (see examples in Figure~\ref{fig:long_horizon_examples}). 
For \projectname-Short, we highlight the target part for the step and distinguish it from the receiving assembly. All CAD-based goals indicate the viewing directions relative to the robot. 
These diagrams specify the part arrangement and spatial relationships needed to complete the assembly.

\begin{figure*}[!ht]
    \centering
    \includegraphics[width=\textwidth]{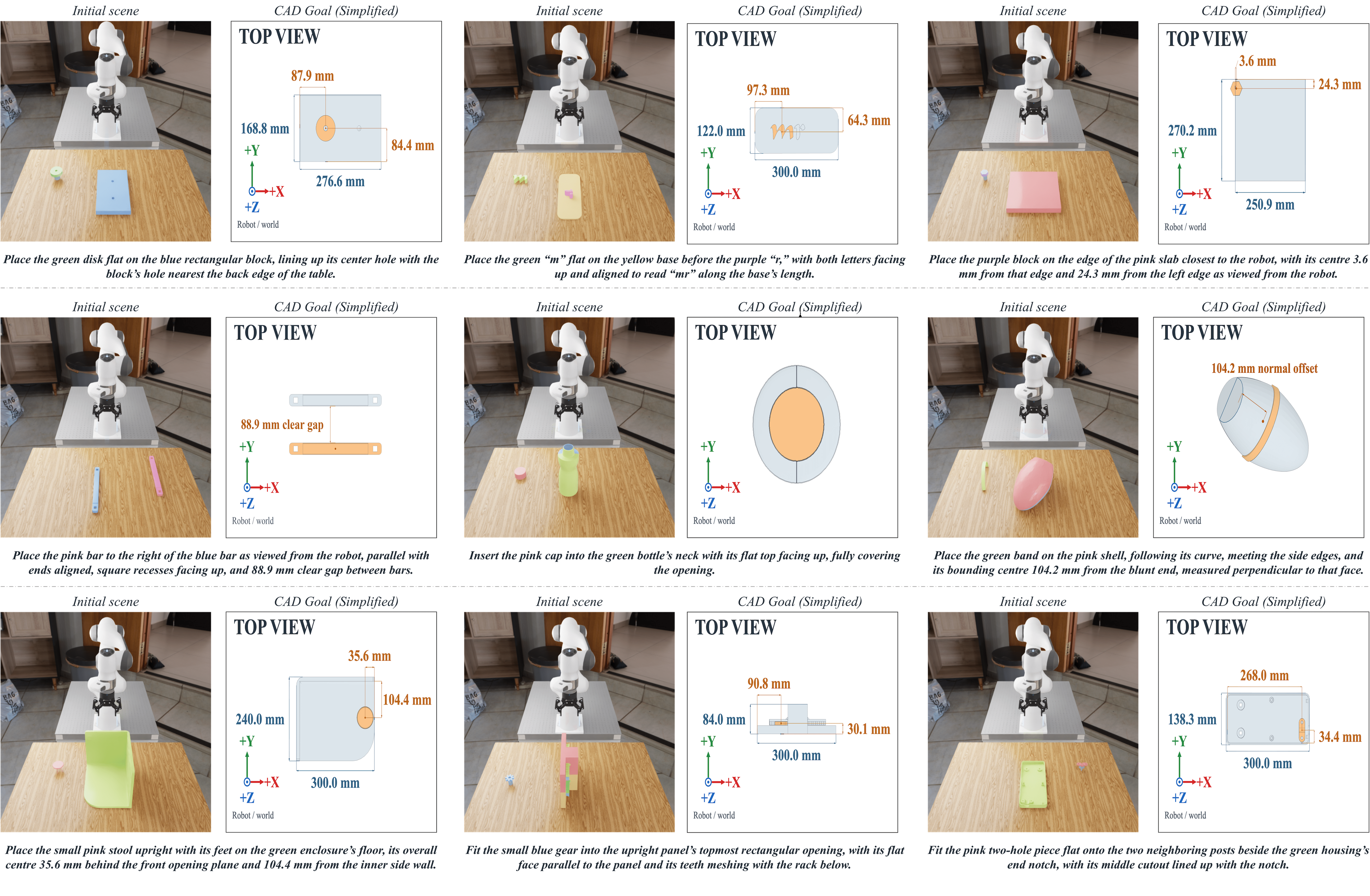}
    \vspace{-20pt}
    \caption{Examples of language and CAD-based goals for tasks in \projectname-Short. Language based goals are constructed with an LLM and manually edited to remove ambiguities. Tasks in \projectname-Short require the robot to use features of the receiving assembly, follow dimensions specified in text or images, or insert parts into narrow spaces. 
    }
    \vspace{-5pt}
    \label{fig:short-horizon-examples}
\end{figure*}

\paragraph{Language goals.}
We use GPT~6.1~Sol to generate language goals for language-based policies for tasks in \projectname-Short. 
The LLM receives the initial scene and the corresponding CAD-based goal and uses the assembly dimensions and relationships to describe the required placement.
We generate five diverse instructions per task to choose from, each identifying the target part, the receiving assembly, and the required placement. 
We manually review and edit these instructions to remove ambiguity and ensure consistency with the target configuration. 
Figure~\ref{fig:short-horizon-examples} shows example tasks.
We do not provide language goals for \projectname-Long, as describing multi-step assemblies in natural language yields long, dense instructions that are poorly suited to specifying precise target configuration.




\subsection{Experimental Setup}
\label{subsec:exp-setup}

\paragraph{Evaluated models and action interfaces.}
We evaluate two VLAs, $\pi_{0.5}$~\cite{pmlr-v305-black25a} and GR00T N1.7~\cite{bjorck2025gr00t}, the WAM Cosmos3-Nano-Policy (Cosmos)~\cite{nvidia2026cosmos3}, and four LLMs: Qwen~3.8~Flash~Next, Opus~5.5 (Opus), Gemini~4~Argon (Argon), and GPT~6~Astra (Astra).
VLA and WAM policy models return fixed chunks of 16 or 32 absolute joint states while LLMs return variable chunks of 3--15 eight-dimensional end-effector actions.
By default, the LLMs receive four RGB camera views: the DROID shoulder-left, shoulder-right, and wrist views, plus an additional front view.
Unless otherwise specified, we use direct inference, in which the LLM maps the four camera views and the goal specification directly to end-effector actions, with medium reasoning effort and a 5M-token budget per task.

\paragraph{Harness configuration.}
We evaluate Astra with the agentic Codex harness in an isolated session, using RGB observations and language or CAD-based goals.
The harness provides file and image tools, episode memory, and an isolated Python environment to write and execute code for numerical calculations and image analysis.
This environment exposes the supplied observations, camera calibration, robot feedback, and task goal, without host-shell or network access, GPU computation, object meshes, ground-truth object poses, hidden target transforms, or evaluator scores.
Robot actions use separate control tools; the harness provides no pretrained perception service or collision-free motion planner.
The token allowance is 10M per task, compared with 5M for direct inference.
This comparison measures the combined effect of harness access and a larger token allowance.

\paragraph{Long-horizon task execution.}
For \projectname-Long, we evaluate Astra with CAD-based goals, four RGB camera views, and the Codex harness, with a cumulative allowance of 100M tokens per assembly, including context compaction.
Each step requires the robot to place one incoming part.
To support assembly with a single arm, a \texttt{merge} tool creates a fixed joint at the achieved relative pose after checking sufficient contact and no penetration. 
The runtime enforces the provided assembly order and requires intermediate parts to be released before the next step; the final part may remain held.
A \texttt{reset\_robot\_to\_init} tool allows nonphysical recovery by resetting the robot while preserving part poses and existing joints.

\subsection{Evaluation Metrics}
\label{subsec:metrics}


\paragraph{Positional accuracy.}
For each short-horizon task $i$, centroid error is $d_i = \|\mathbf{c}_i-\mathbf{c}_i^\star\|_2$ in millimeters, where $\mathbf{c}_i$ and $\mathbf{c}_i^\star$ are the achieved and target centers of mass under uniform density.
Across $N$ episodes, we report positional success rate as $\operatorname{SR}@\tau = \frac{100}{N}\sum_{i=1}^{N}\mathbf{1}[d_i \leq \tau]$ for $\tau \in \{1,2,5,10\}$\,mm.
We report mean and median centroid error across all tasks, including unsuccessful episodes, with a 2\,m cap because the simulator has no floor and dropped objects can continue falling indefinitely.
SR and 3D IoU use uncapped metrics.

\paragraph{Volumetric overlap.}
For the assembly part's achieved and target volumes $V_i$ and $V_i^\star$, 3D intersection over union (3D IoU) measures overlap.
We compute $\operatorname{IoU}_i = 100\,\operatorname{Vol}(V_i \cap V_i^\star)/\operatorname{Vol}(V_i \cup V_i^\star)$ and report the mean across tasks.
We exclude the fixed receiving assembly. 
Unlike centroid error alone, 3D IoU reflects both translation and orientation, subject to the part's symmetries.

\paragraph{Assembly progress and placement fidelity.}
For \projectname-Long, each step places one incoming part.
A successful \texttt{merge} completes a step. We report completed steps relative to required placements and evaluate placement accuracy separately.
Parts@$\tau$ counts completed steps whose parts have terminal centroid error at most $\tau$\,mm.
Whole-assembly positional success requires every required incoming part to meet the same tolerance. The moving part may remain grasped at termination, including for angled target configurations that require support. Our evaluation measure achieved placement fidelity and while post-release stability checks are out of scope.

We score part poses at the final recorded physics tick in the world frame, with fixed part identities and no permutation matching.
We report mean per-part 3D IoU, its standard deviation (SD), mean and median centroid error under the same 2\,m cap used for \projectname-Short.
These summaries include all required incoming parts, including unplaced parts, and exclude the initial base.
Overall statistics use all 65 required parts (across 15 tasks in \projectname-Long), without averaging assembly-level summaries.

\paragraph{Runtime.}
Episode wall time includes model and tool execution, rather than robot-motion time alone.
We report mean $\pm$ sample SD across episodes.
For \projectname-Long, runtime SD describes variation across assemblies, while 3D IoU SD describes variation across required parts.

\section{Results}
\label{sec:results}

\begin{table*}[!t]
\centering\small
\setlength{\tabcolsep}{3pt}
\renewcommand{\arraystretch}{1.0}
\caption{\projectname-Short: 280 tasks per row.
IoU/SR are percentages; SR thresholds are in millimeters.
Mean and median centroid errors use a 2\,m cap.
Time is mean $\pm$ SD in minutes.
L: language/direct; C: CAD-goal/direct; H: language/Codex harness;
CH: CAD-goal/Codex harness.
Direct/harness limits are 5M/10M tokens.
Green marks the best and second-best placement scores within each block.}
\label{tab:mainresults}
\label{tab:goal_specification}
\label{tab:runtime_success}
\begin{adjustbox}{max width=\textwidth}
\begin{tabular}{@{}llrrrrrrrr@{}}
\toprule
\textbf{Model} & \textbf{Cond.} & \textbf{3D IoU} & \textbf{SR@1} & \textbf{SR@2} & \textbf{SR@5} & \textbf{SR@10} & \shortstack{\textbf{Mean error}\\\textbf{(mm)}} & \shortstack{\textbf{Median error}\\\textbf{(mm)}} & \shortstack{\textbf{Time}\\\textbf{(min)}} \\
\midrule
$\pi_{0.5}$ (2 views) & L & 0.5 & 0.0 & 0.0 & 0.0 & 0.7 & 425.7 & 301.5 & $9.7 \pm 0.9$ \\
GR00T N1.7 (2 views) & L & 0.2 & 0.0 & 0.0 & 0.0 & 0.0 & 452.2 & 348.3 & $13.5 \pm 1.5$ \\
Cosmos (3 views) & L & 2.6 & 0.0 & 0.0 & 1.1 & 1.4 & 370.2 & 231.0 & $8.7 \pm 0.7$ \\
Qwen 3.8 Flash Next & L & 1.6 & 0.0 & 0.0 & 0.0 & 1.1 & 294.7 & 278.4 & $12.4 \pm 6.3$ \\
Opus 5.5 & L & 15.4 & \secondscore{1.4} & 3.2 & 6.8 & 19.6 & \secondscore{116.1} & 27.0 & $21.3 \pm 7.3$ \\
Argon 4 & L & \secondscore{18.2} & \bestscore{1.8} & \secondscore{5.0} & \secondscore{12.9} & \secondscore{26.8} & 134.9 & \secondscore{23.7} & $24.6 \pm 14.6$ \\
GPT 6 Astra & L & \bestscore{29.6} & \bestscore{1.8} & \bestscore{6.8} & \bestscore{25.7} & \bestscore{48.2} & \bestscore{74.8} & \bestscore{10.3} & $7.9 \pm 3.2$ \\
\midrule\multicolumn{10}{@{}l}{\textit{Additional goal/harness setups (same 280 tasks)}} \\
Opus 5.5 & C & 16.8 & 0.7 & 1.8 & 7.9 & 22.1 & 101.2 & 24.5 & $19.3 \pm 10.4$ \\
Argon 4 & C & 20.2 & 1.8 & 3.9 & 14.3 & 31.4 & 107.5 & 18.2 & $21.8 \pm 13.1$ \\
GPT 6 Astra & C & 35.1 & \secondscore{4.6} & 14.3 & 32.5 & 52.9 & \bestscore{71.7} & 8.8 & $7.6 \pm 3.3$ \\
GPT 6 Astra & H & \secondscore{40.4} & \bestscore{5.7} & \secondscore{15.7} & \secondscore{44.3} & \secondscore{60.7} & 104.2 & \secondscore{6.3} & $8.7 \pm 4.4$ \\
GPT 6 Astra & CH & \bestscore{42.7} & 2.5 & \bestscore{16.4} & \bestscore{51.8} & \bestscore{69.3} & \secondscore{90.6} & \bestscore{4.9} & $8.2 \pm 3.8$ \\
\bottomrule
\end{tabular}
\end{adjustbox}
\vspace{-10pt}
\end{table*}

We evaluate whether current RFMs can place parts precisely, whether CAD-based goals or an agentic harness improve placement, and how accurately can the strongest model complete multi-step assemblies.
We first compare models and interventions on the 280 tasks in \projectname-Short, then evaluate the strongest model (Astra) on 15 multi-step assemblies in \projectname-Long.

\subsection{Placement precision remains limited}
\label{sec:short_results}

Table~\ref{tab:mainresults} compares all RFMs with language goals.
Astra, the strongest model in this setting, achieves 48.2\% SR@10\,mm but drops to 1.8\% SR@1\,mm.
It achieves the highest mean 3D IoU and the highest or joint-highest positional success rate at every threshold.
Opus, Argon, and Astra outperform the evaluated VLAs and WAM on mean 3D IoU and positional success rates, but precise placement across all models remains difficult (Figure~\ref{fig:centroid_error_threshold}). Ablations on the number of camera inputs to LLMs are in the supplementary material.

\begin{figure}[!ht]
    \centering
    \includegraphics[width=\columnwidth]{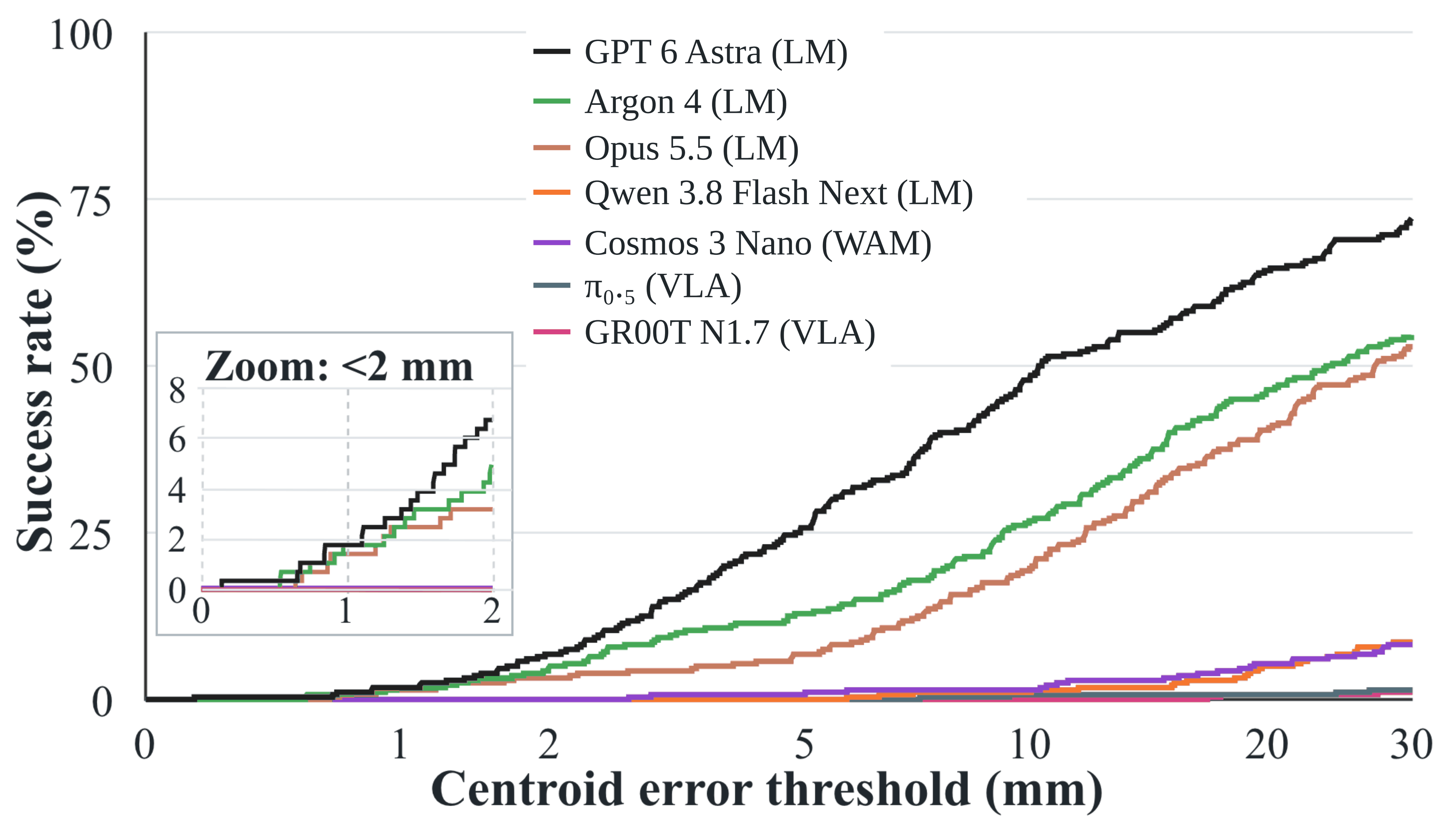}
    \caption{
    Success rate of evaluated RFMs on \projectname-Short for centroid error threshold up to 30mm.   
    Success rises with looser thresholds, reaching 72\% for Astra at 30\,mm, but falls sharply at millimeter scale: no model exceeds 7\% within 2\,mm (inset).
    }
    \vspace{-20pt}
    \label{fig:centroid_error_threshold}

\end{figure}

\paragraph{LLM control takes minutes.}
Mean episode wall time ranges from 7.9 to 24.6 minutes across directly prompted LLMs.
Astra and Argon have comparable performance at the strictest positional tolerances, but their mean episode wall times differ substantially: 7.9 versus 24.6 minutes.

\subsection{CAD-based goals and harness access improve placement}
\label{sec:cad_results}
\label{sec:harness_results}

\paragraph{CAD-based goals.}
Replacing language with CAD-based goals improves mean 3D IoU, SR@5\,mm, and SR@10\,mm for Opus, Argon, and Astra (Table~\ref{tab:mainresults}).
For Astra, mean 3D IoU increases from 29.6\% to 35.1\%, and SR@2\,mm increases from 6.8\% to 14.3\%.
Gains vary with model and tolerance: Opus's SR@2\,mm decreases from 3.2\% to 1.8\%.

\paragraph{Codex harness.}
With language goals, Astra with the Codex harness increases mean 3D IoU from 29.6\% to 40.4\%, SR@1\,mm from 1.8\% to 5.7\%, and SR@10\,mm from 48.2\% to 60.7\%.
Median centroid error decreases from 10.3 to 6.3\,mm.
Mean reported token use increases from 2.13M to 3.13M per episode (+47.0\%), while mean episode wall time increases from 7.89 to 8.68 minutes (+9.9\%).
This configuration also raises the token allowance from 5M to 10M per task.
%
Both interventions improve placement, but substantial room for improvement remains: even the strongest configuration reaches only 5.7\% SR@1\,mm and 60.7\% SR@10\,mm.
Astra with CAD-goal and codex harness achieves the highest 3D IoU of 42.7\%, and outscores other baselines and configurations on 2/5/10mm success rate thresholds. 

\vspace{-10pt}

\paragraph{Observed control strategies. }Astra uses codex sessions to write custom python tools for spatial estimation. Common workflows combined calibrated multiview triangulation, ray–plane intersections, reprojection checks, and rigid-body      transformations to estimate feature locations and account for object-to-gripper offsets. Astra  
 also inspected cropped image regions, revised geometric hypotheses after new observations, and      
 occasionally generated Cartesian waypoint sequences using orientation interpolation.

\subsection{Precise long assemblies are challenging}
\label{sec:long_results}
\begin{table*}[!t]
\centering
\small
\setlength{\tabcolsep}{3pt}
\renewcommand{\arraystretch}{1.0}
\caption{\projectname-Long: Astra with projected CAD goals and the Codex harness on 15 assemblies.
Steps completed counts successful merges, parts getting attached. Parts@$\tau$ counts those within $\tau$\,mm.
IoU is mean $\pm$ SD across all required parts, including unplaced parts while the base is excluded.
Centroid errors use a 2\,m cap.
The overall row summarizes all 65 parts; time is mean $\pm$ SD across 15 assemblies.
Green marks the best and second-best IoU and error scores.}
\vspace{-10pt}
\label{tab:long_results}
\begin{adjustbox}{max width=\textwidth}
\begin{tabular}{@{}lc*{4}{>{\centering\arraybackslash}p{1.2em}}rrrr@{}}
\toprule
\textbf{Assembly} & \shortstack{\textbf{Steps}\\\textbf{completed}} & \multicolumn{4}{c}{\textbf{Parts@ (mm)}} & \textbf{3D IoU (\%)} & \shortstack{\textbf{Mean error}\\\textbf{(mm)}} & \shortstack{\textbf{Median error}\\\textbf{(mm)}} & \shortstack{\textbf{Time}\\\textbf{(min)}} \\
\cmidrule(lr){3-6}
& & \textbf{1} & \textbf{2} & \textbf{5} & \textbf{10} & & & & \\
\midrule
Kyrre seat & 3/3 & 0 & 0 & 0 & 1 & 15.1 $\pm$ 12.8 & 13.3 & 12.7 & 155.1 \\
Mammut table & 4/4 & 0 & 0 & 3 & 4 & \bestscore{72.9 $\pm$ 6.4} & \bestscore{4.1} & \secondscore{4.0} & 57.9 \\
Froset chair & 1/3 & 0 & 0 & 0 & 0 & 4.6 $\pm$ 8.0 & 234.2 & 78.4 & 70.6 \\
Omtanksam chair & 4/4 & 0 & 0 & 0 & 1 & 4.6 $\pm$ 9.2 & 102.7 & 113.9 & 118.3 \\
Beam assembly & 4/4 & 0 & 0 & 0 & 0 & 0.0 $\pm$ 0.0 & 193.6 & 194.7 & 111.9 \\
Jokkmokk table & 0/4 & 0 & 0 & 0 & 0 & 0.0 $\pm$ 0.0 & 429.8 & 409.8 & 60.8 \\
Lack table & 0/9 & 0 & 0 & 0 & 0 & 0.0 $\pm$ 0.0 & 576.3 & 480.2 & 10.9 \\
Pinnig shelf & 2/6 & 0 & 0 & 0 & 0 & 0.0 $\pm$ 0.0 & 458.6 & 315.5 & 62.6 \\
Nilsolle seat & 2/5 & 0 & 0 & 0 & 0 & 0.1 $\pm$ 0.2 & 183.1 & 149.5 & 264.6 \\
Lisabo chair & 1/9 & 0 & 0 & 0 & 0 & 0.0 $\pm$ 0.0 & 631.0 & 543.3 & 105.9 \\
Stefan chair & 2/2 & 0 & 0 & 0 & 0 & 10.1 $\pm$ 6.3 & 16.2 & 16.2 & 125.9 \\
Bernhard chair & 1/2 & 0 & 0 & 0 & 1 & 8.6 $\pm$ 12.1 & 331.2 & 331.2 & 21.2 \\
Perjohan stool & 0/3 & 0 & 0 & 0 & 0 & 0.0 $\pm$ 0.0 & 433.9 & 499.8 & 178.7 \\
Sjalland chair & 3/3 & 0 & 0 & 0 & 2 & 8.4 $\pm$ 14.0 & 15.8 & 9.2 & 165.6 \\
Nils chair & 4/4 & 0 & 1 & 3 & 3 & \secondscore{66.0 $\pm$ 6.6} & \secondscore{4.5} & \bestscore{2.9} & 92.1 \\
\midrule
Overall / mean & 31/65 & 0 & 1 & 6 & 11 & $10.1 \pm 23.1$ & 315.2 & 209.1 & $106.8 \pm 65.9$ \\
\bottomrule
\end{tabular}
\end{adjustbox}
\end{table*}

\begin{figure*}[!t]
    \centering
    \includegraphics[width=\textwidth]{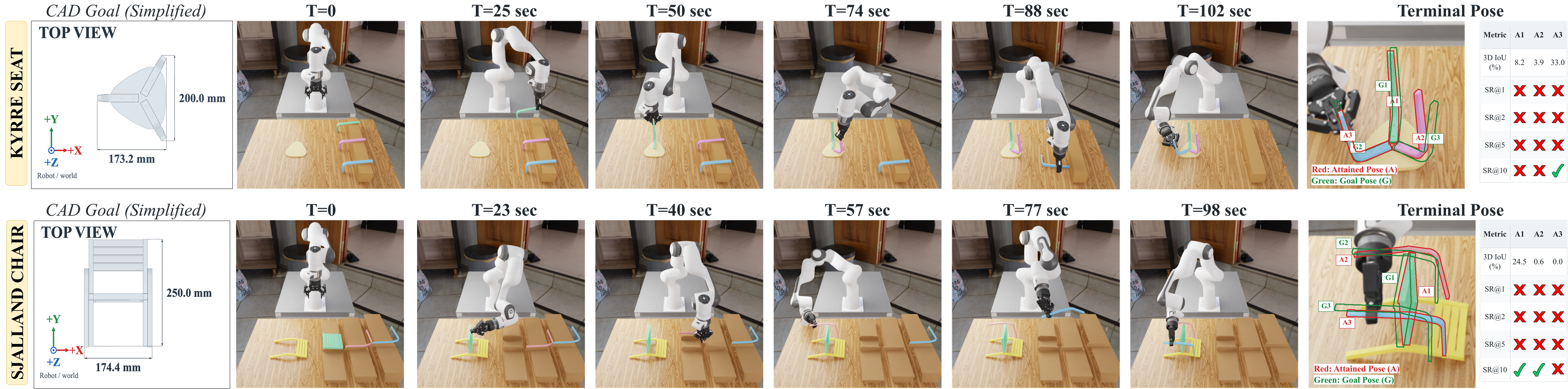}
    \caption{Example rollouts from \projectname-Long using Astra with CAD-based goals and the Codex harness. Completed steps indicate assembly progress; terminal placement accuracy is evaluated separately.}
    \vspace{-5pt}
    \label{fig:long-horizon-rollouts}
\end{figure*}


As shown in Table~\ref{tab:long_results}, Astra completes all required steps in seven of the 15 \projectname-Long assemblies. 
Only the \textit{Mammut} table assembly meets the 10 mm centroid-error threshold for each required part. 
No assembly meets the 1, 2, or 5\,mm tolerance for any required part.
These results separate assembly progress from placement fidelity: completion of a step does not ensure that its part reaches the target pose (Figure~\ref{fig:long-horizon-rollouts}).

Across all 15 assemblies, Astra completes 31 of 65 steps (47.7\%).
Of these completed steps, 0, 1, 6, and 11 place their parts within 1, 2, 5, and 10\,mm, respectively.
Across all 65 required parts, mean 3D IoU is $10.1 \pm 23.1\%$, and capped mean and median centroid errors are 315.2 and 209.1\,mm.
Mean episode wall time is $106.8 \pm 65.9$\,min.
These results highlight the need for methods that maintain reliable placement accuracy across long assembly sequences.

\subsection{Real World Validation}

\begin{figure*}[!t]
    \centering
    \includegraphics[width=\textwidth]{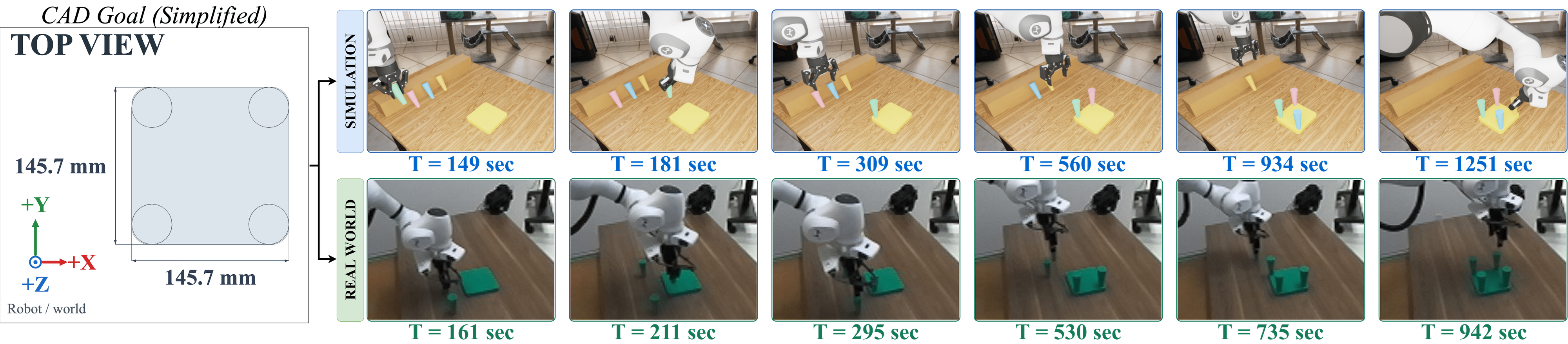}
    \caption{Comparison showing simulation and real rollouts with the same CAD-based goal input and robot setup. The robot policy used in both cases is Astra with a codex harness.
    \vspace{-15pt}
    }
    
    \label{fig:sim-real-comparison}
\end{figure*}

We chose the \textit{Mammut} table and \textit{Beam} assembly tasks to 3D print and conduct qualitative studies of the performance of GPT~6~Astra in the real world. 
The model was used with the same configuration as in \projectname-Long (with codex harness, CAD-based goal input and a 100M token budget). 
Astra was incapable of assembling the \textit{Beam}.
Figure~\ref{fig:sim-real-comparison} shows a side-by-side comparison of the rollout between simulation and real world for the \textit{Mammut} task. 
Quantitative results show that placing all pegs did not ensure symmetric placement or the correct distances from the seat edges.
While recent methods have advanced
6D object pose estimation and tracking~\cite{foundationpose},
reliably measuring the poses of assembled parts and evaluating
their placement fidelity at millimeter-scale tolerances remain
challenging.

\section{Conclusion}
\label{sec:conclusion}

\vspace{-2mm}
Precise assembly requires accurate placement and reliable verification of the final result. 
RoboBuilder measures this accuracy through centroid error and 3D IoU across short- and long-horizon tasks.
Our results show that current robot foundation models remain far from reliable assembly, despite improvements from CAD-based goals and computational tools.
Extending this evaluation to physical systems raises an open problem in pose estimation. Reliable completion measurements require accurate estimates of part position and rotation, including when assembled parts are partly hidden.

\section*{LLM Usage Statement}

GPT 6.1 Sol was used to generate textual goal instructions for tasks in \projectname-Short, which then was verified with manual reviews. No LLM was used to create assets or assemblies. Authors used Opus 5.5 and GPT 6 Astra to proof read the text, to search literature and to write and debug code for simulation design and evaluations. The LLMs evaluated in this paper were called through their official endpoints and had no access to simulator internals or privileged ground truth. The authors checked all experiments, numbers, and claims and take full responsibility for the paper.
{
    \small
    \bibliographystyle{ieeenat_fullname}
    \nocite{kimble2020nist}
    \bibliography{main}
}


\end{document}